\documentclass{ifacconf}
\usepackage{natbib}        
\usepackage{graphicx}      
\usepackage{natbib}        
\usepackage{amsmath, amssymb}
\usepackage{xcolor}
\usepackage{subcaption} 

\usepackage{algorithmic,algorithm}
\DeclareMathOperator*{\Argmin}{Argmin}
\begin{document}
\begin{frontmatter}

\title{Curriculum-Learned Vanishing Stacked Residual PINNs for\\Hyperbolic PDE State Reconstruction\thanksref{footnoteinfo}} 

\thanks[footnoteinfo]{Sponsor and financial support acknowledgment
goes here. Paper titles should be written in uppercase and lowercase
letters, not all uppercase.}

\author[First]{Katayoun Eshkofti} 
\author[First]{Matthieu Barreau} 
            
\address[First]{Digital Futures and KTH Royal Institute of Technology, Stockholm, Sweden (e-mail: {eshkofti,barreau}@kth.se).}

\begin{abstract}                
Modeling distributed dynamical systems governed by hyperbolic partial differential equations (PDEs) remains challenging due to discontinuities and shocks that hinder the convergence of traditional physics-informed neural networks (PINNs). The recently proposed vanishing stacked residual PINN (VSR-PINN) embeds a vanishing-viscosity mechanism within stacked residual refinements to enable a smooth transition from the parabolic to hyperbolic regime. This paper integrates three curriculum-learning methods as primal-dual (PD) optimization, causality progression, and adaptive sampling into the VSR-PINN. The PD strategy balances physics and data losses, the causality scheme unlocks deeper stacks by respecting temporal and gradient evolution, and adaptive sampling targets high residuals. Numerical experiments on traffic reconstruction confirm that enforcing causality systematically reduces the median point-wise MSE and its variability across runs, yielding improvements of nearly one order of magnitude over non-causal training in both the baseline and PD variants.
\end{abstract}


\end{frontmatter}

\section{Introduction}

Quasilinear hyperbolic partial differential equations (PDEs) form a fundamental class of models for distributed parameter systems, particularly in fields such as traffic flow, fluid dynamics, and networked infrastructure. Their mathematical structure captures the propagation of waves, shocks, and other transport phenomena, making them essential for real-time control and state estimation tasks. In traffic systems, these PDEs underpin both classical and modern approaches to traffic reconstruction and identification, where the objective is to infer traffic states from limited and noisy measurements.

The mathematical richness of quasilinear hyperbolic PDEs provides a robust foundation for control design but also poses significant analytical and numerical challenges, particularly in the presence of non-smooth and shock-dominated regimes \citep{Li_2010}. Although advanced boundary control and observer techniques have substantially extended the practical applicability of quasilinear hyperbolic PDEs in control \citep{Deutscher_feedback, Kitsos_Lyapunov}, obstacles remain in addressing nonlinearities, time delays, and networked topologies.

The shift from purely model-based methods to hybrid and learning-based approaches reflects the growing need for physical fidelity and adaptability to real-world data limitations. Advanced numerical methods often struggle to maintain accuracy, whereas learning-based approaches offer enhanced estimation accuracy and robustness \citep{Cassia2025, Shi2022}. In this direction, \citep{morand2024} showed that neural networks can learn conservative first-order finite-volume solvers by optimizing numerical flux, improving forward prediction accuracy for nonlinear conservation laws while still preserving mass conservation.

Among these, physics-informed neural networks (PINNs), initially proposed by \citep{raissi2019physics}, represent a powerful framework for solving PDEs by minimizing a loss function that incorporates the residuals of the governing equations, along with initial and boundary conditions, and data constraints. However, baseline PINNs face limitations due to their inability to capture discontinuities, which leads to instability and loss of accuracy near shocks \citep{DeRyck2024}.

To overcome these limitations, recent advances in deep learning architectures, particularly stacked \citep{AmandaStacked} and residual networks \citep{Chen2023} have enhanced the representational capacity of PINNs. However, these developments alone remain insufficient for robust state reconstruction in strongly nonlinear hyperbolic systems. This limitation has motivated the adoption of curriculum learning strategies, which structure the training process to gradually increase task complexity, thereby improving convergence and generalization \citep{wang_causality}.

Building upon this direction, the vanishing stacked residual PINN (VSR-PINN), proposed by \citep{eshkofti2025}, introduces a stacked residual architecture composed of multiple interconnected neural blocks with adaptive residual connections to enhance the reconstruction of hyperbolic PDEs. Inspired by the artificial viscosity concepts in computational fluid dynamics, a viscosity-decreasing mechanism is employed to stabilize training \citep{He_viscosity, Caldana2025}. By progressively reducing viscosity across stacks in the form of a functional series, the network learns to accurately resolve shocks and discontinuities \citep{eshkofti2025}. Additionally, this approach optimizes the loss functions of all stacks simultaneously while assigning a unit weight to the physics-based loss term. Related work, such as incremental PINN \citep{Dekhovich_ipinn}, has explored overlapping subnetworks for the sequential learning of related PDEs, thereby improving generalization and performance.
Balancing the smoothing effect of viscosity with the need for sharp shock resolution requires the implementation of adaptive control mechanisms. To this end, curriculum extensions such as loss weighting \citep{McClenny2023, barreau_accuracy}, causality-driven progression \citep{wang_multistage, Mattey_sequential}, and adaptive sampling \citep{Lu_deepxde, Mao_adaptive, WU_adaptive} have shown promise in enhancing convergence without degrading physical fidelity.

This paper advances this line of research by integrating three complementary curriculum-based strategies within the vanishing-viscosity framework. Primal–Dual (PD) optimization is employed to improve loss balancing and constraint handling, temporal and stack-wise causality introduces an unlocking mechanism for progressive learning, and adaptive sampling directs computational focus to regions with high residuals. While these methods have been developed independently, their joint integration into a single PINN architecture remains novel and largely unexplored. This combined approach, embedded within a vanishing-viscosity curriculum framework, aims to enhance training efficiency, robustness, and physical consistency.

The remainder of paper is organized as follows: Section 2 provides the mathematical background and formulates the governing quasilinear hyperbolic PDE, together with the vanishing viscosity method and its integration into the VSR-PINN framework. Section 3 presents the three curriculum learning methods adapted to the VSR PINN architecture. Section 4 describes the numerical implementation and evaluates the performance of the proposed methods on the traffic reconstruction problem. Finally, Section 5 summarizes the main findings and outlines potential future directions.

\paragraph*{Notations:}
The set of real numbers is denoted by $\mathbb{R}$, while $\mathbb{R}^{+}$ signifies the set of nonnegative real numbers. For single-variable functions, derivatives are represented by prime symbol $f'$. In the case of multivariate functions, partial derivatives with respect to space and time are indicated by $\partial f$ with appropriate subscripts. The space $C^{1}(\mathbb{R}^{+} \times \mathbb{R})$ refers to functions that are continuously differentiable on the given domain. Throughout this work, $L^p(\Lambda)$, $\mathcal{H}^k(\Lambda)$, and $L^{\infty}(\Lambda)$ shall denote the Lebesgue, Sobolev, and essentially bounded function spaces on $\Lambda$, respectively. 

\section{Background and problem formulation}

Let $\Lambda := [0, T] \times [0, L] \subset \mathbb{R}^{+} \times \mathbb{R}$ be a bounded spatiotemporal domain and $\mathcal{H}$ is an appropriate Sobolev space with $\mathcal{H} = L^2(0,T;H^1(0,L))\cap H^1(0,T;H^{-1}(0,L))$. The following one-dimensional quasilinear hyperbolic PDE is considered:
\begin{equation}
    \begin{cases}
    \label{eq:system}
r[u](t, x) = \partial_{t} u + \partial_{x} f(u)  & (t,x) \in \Lambda, \\[6pt]
u(0, x) = u^{0}(x), &  x \in [0,L], \\[6pt]
u(t, 0) = u_{b}^{-}(t), \quad u(t, L) = u_{b}^{+}(t), & t \in [0,T],
\end{cases}
\end{equation}
where $u \in \mathcal{H}$ is the PDE solution, and $r := \mathcal{H} \rightarrow \mathcal{L}^{2}(\Lambda, \mathbb{R})$ denotes the nonlinear PDE residual operator. The initial datum $u^0:[0,L]\to\mathbb{R}$ is of bounded variation. The flux function $f\in C^2(\mathbb{R})$, so the characteristic speed $a(u):=f'(u)$ is continuous with real values, and the scalar conservation law is hyperbolic. To control growth and ensure $r[u]\in \mathcal{L}^2(\Lambda)$ for $u\in\mathcal{H}$, it is assumed that $f$ is globally Lipschitz on the range attained by $u$. The boundary data $u_b^{-},u_b^{+}:[0,T]\to\mathbb{R}$ specify, in the sense of characteristics, only the incoming components at $x=0$ and $x=L$, respectively; corner compatibility is imposed, $u^0(0)=u_b^{-}(0)$ and $u^0(L)=u_b^{+}(0)$. 

Hyperbolic equations often exhibit sharp discontinuities, therefore vanishing viscosity method can be employed which introduces an artificial viscosity to smooth shocks and regularize solutions. Therefore the parabolic relaxation of ~\eqref{eq:system} is defined as
\begin{equation}\label{eq:parabolic}
    \partial_t u_{\gamma} + \partial_x f(u_{\gamma}) = \gamma \partial_{xx} u_{\gamma}, \quad \gamma > 0,
\end{equation}
where $\gamma$ is small positive coefficient and $u_{\gamma}$ is the unique smooth solution corresponding to that $\gamma$. Energy calculations yield uniform in $\gamma$ bounds 
and an entropy inequality with an $\mathcal{O}(\gamma)$ remainder. These bounds provide compactness, so along a 
\( u_{\gamma} \to u \) as \( \gamma \to 0^+ \).  
Passing to the limit removes the viscous term and preserves the entropy inequality; 
hence, the limit \( u \) satisfies both the conservation law in the weak sense 
and the entropy admissibility.


Initially, the goal is to approximate the solution \( u^{(0)} \) of the parabolic PDE~\eqref{eq:parabolic}, which is well-posed and admits a smooth solution for a fixed initial viscosity coefficient \( \gamma_{\text{init}} > 0 \). A standard feedforward neural network \( \hat{u}^{(0)}(\cdot; \boldsymbol{\theta}_0) \) with \( L \) hidden layers is expressed as
\[
\hat{u}^{(0)}(t, x; \boldsymbol{\theta}_0) = W_L H_{L-1} \circ \cdots \circ H_1(t, \mathbf{x}) + b_L 
\triangleq \mathcal{N}([t, x]; \boldsymbol{\theta}_0),
\]
where the network takes spatiotemporal coordinates \( (t, x) \), such that \( (t, x) \in \Lambda \).  
The superscript \( (\cdot)^{(0)} \) indicates the initial stage, which is the baseline PINN. For each hidden layer indexed by \( l = 1, \ldots, L - 1 \), the corresponding feature map is defined as $H_l(\nu) = \phi(W_l \nu + b_l)$, where \( W_l \in \mathbb{R}^{n_l \times n_{l-1}} \) and \( b_l \in \mathbb{R}^{n_l} \) denote the weight matrix and bias vector of the \( l^{\text{th}} \) layer, respectively. The complete set of trainable network parameters is denoted by
\[
\boldsymbol{\theta}_0 = \{ W_l, b_l \}_{l=1}^L,
\]
and \( \phi \in C^{\infty}(\mathbb{R}, \mathbb{R}) \) represents a smooth, element-wise activation function applied at each layer. Formally, the baseline network $\hat{u}^{(0)}(\cdot;\boldsymbol{\theta}_0)$ is trained by solving
\begin{equation}
   \boldsymbol{\theta}_0^* = \begin{array}[t]{cl}
        \Argmin_{\boldsymbol{\theta}} 
\int_{\Gamma} 
\| u(\nu) - \hat{u}^{(0)}(\nu; \boldsymbol{\theta}_0) \|^2 d\nu
\\ \text{s.t.} \quad
\int_{\Lambda} 
\left| r_{\gamma_{\text{init}}}(\cdot; \hat{u}^{(0)}(\cdot; \boldsymbol{\theta}_0)) \right|^2 = 0.
\end{array} 
\label{eq:training}
\end{equation}

where $\Gamma = \Gamma_{\text{init}} \cup \Gamma_{\text{boundary}}$ collects the points where initial and boundary conditions are specified and $r_{\gamma_{\text{init}}}$ is the viscous PDE residual defined as 
\[
r_{\gamma_{\text{init}}}(\cdot; \hat{u}^{(0)}) = \partial_t \hat{u}^{(0)} + \partial_x f(\hat{u}^{(0)}) - \gamma_{\text{init}} \, \partial_{xx} \hat{u}^{(0)}.
\]
Since the integrals are intractable, Monte Carlo sampling and Lagrangian relaxation are employed, and the following surrogate problem is optimized:
\[
\boldsymbol{\theta}_0^* = \operatorname*{Argmin}_{\boldsymbol{\theta}} \max_{\lambda > 0} 
\mathcal{L}_{\lambda}(\hat{u}^{(0)}, \gamma_{\text{init}}),
\]
where \( \mathcal{L}_{\lambda}(\hat{u}^{(0)}, \gamma_{\text{init}}) \) denotes the total loss function with viscosity $\gamma_{\text{init}}$, expressed as
\begin{equation*}
    \mathcal{L}_{\lambda}(\hat{u}^{(0)}, \gamma, \mathcal{D}_{\text{phy}}) = \\ \mathcal{L}_{\text{data}}(\hat{u}^{(0)}) + \lambda \, \mathcal{L}_{\text{phy}}(\hat{u}^{(0)},  \gamma, \mathcal{D}_{\text{phy}}),
\end{equation*}
with
\[
\mathcal{L}_{\text{data}}(\hat{u}^{(0)}) = 
\frac{1}{|\mathcal{D}_{\text{data}}|} 
\sum_{(t_k, x_k, u_k) \in \mathcal{D}_{\text{data}}} 
|u_k - \hat{u}^{(0)}(t_k, x_k)|^2,\]
\[
\mathcal{L}_{\text{phy}}(\hat{u}^{(0)}, \gamma, \mathcal{D}_{\text{phy}}) = 
\frac{1}{|\mathcal{D}_{\text{phy}}|} 
\sum_{(t_k, x_k) \in \mathcal{D}_{\text{phy}}} 
|r_{\gamma}(t_k, x_k; \hat{u}^{(0)})|^2.
\]
Here, these losses are defined over discrete sets 
\( \mathcal{D}_{\text{data}} \subset \Gamma \) and 
\( \mathcal{D}_{\text{phy}} \subset \Lambda \).  
In this formulation, the Lagrangian multiplier is considered fixed. However, in the next section, the PD update method is introduced, which automatically adjusts this coefficient. Additionally, $(t_k, x_k)$ pair represents spatiotemporal points at iteration $k$. 
This minimization problem 
$\mathcal{L}_{\lambda}(\hat{u}^{(0)}, \gamma, \mathcal{D}_{\text{phy}}) $
yields a stable low-fidelity estimator consistent with the parabolic regularization.

\paragraph*{VS-PINN} As proposed by \citep{eshkofti2025}, this vanishing viscosity approach can be integrated in stacked residual PINN. The first component of the architecture is the baseline PINN \citep{raissi2019physics} which functions as the parabolic regularization. The goal is to approximate the entropy-admissible solution $u$ of PDE \eqref{eq:system} by following the vanishing viscosity path, which transitions from a parabolic surrogate equation~\eqref{eq:parabolic} to the desired hyperbolic PDE~\eqref{eq:system}.

To achieve this, the proposed method couples a viscous base PINN with a series of stacked residual PINNs. The viscosity coefficients in these stacked networks gradually vanish across the stacks, guiding the model from a smooth viscous approximation toward the physically consistent, entropy-admissible hyperbolic solution. 

To progressively sharpen the solution, \( \hat{u}^{(0)} \) is refined using \( n \) stacked residual networks.  
At stack \( i \in \{1, \ldots, n\} \),
\begin{multline} \label{eq:stackedrescorrection}
\hat{u}^{(i)}(t, \mathbf{x}; \boldsymbol{\theta}_i) =
\hat{u}^{(i-1)}(t, \mathbf{x}; \boldsymbol{\theta}_{i-1}) + \\
|\alpha_i| \, 
\mathcal{N}\left([t, \mathbf{x}, \hat{u}^{(i-1)}(t, \mathbf{x})]; \boldsymbol{\theta}_i \right),
\end{multline}
where the learnable scalar \( |\alpha_i| \) scales the contribution of block \( i \) and enforces small, stable corrections near the parabolic solution, while enabling shock sharpening as viscosity decreases.  
Each residual block receives the previous prediction \( \hat{u}^{(i-1)} \) as an input feature, acting similarly to an observer that feeds back the model discrepancy. The viscosity \( \gamma_i \) is assigned to the physics loss of stack \( i \), where it approaches zero through a superlinear schedule defined as
\begin{equation}
\label{eq:viscosity}
\gamma_i = \gamma_{\text{init}}
\left[ 1 - \left( \frac{i}{n} \right)^p \right],
\quad p \in (0,1].
\end{equation}
Hence, \( \gamma_0 = \gamma_{\text{init}} > 0 \) and \( \gamma_n = 0 \).  
This smooth-to-sharp curriculum preserves the numerical stability of the early stacks and progressively exposes the network to the discontinuous hyperbolic regime.
The stacked model is trained by minimizing the average stack loss with a small penalty on the residual gains:
\[
\min_{\{\boldsymbol{\theta}_i, \alpha_{i+1}\}_{i=0}^n} 
\mathcal{L}\left(\{\boldsymbol{\theta}_i\}_{i=0}^n\right)
+ \sum_{i=0}^n \alpha_{i+1}^2,\]
\[\mathcal{L}\left(\{\boldsymbol{\theta}_i\}_{i=0}^n\right)
= \frac{1}{n + 1} \sum_{i=0}^n 
\mathcal{L}_{\lambda}\left(\hat{u}^{(i)}, \gamma_i\right).\]

Here one can denote $\Theta={\{\boldsymbol{\theta}_i\}_{i=0}^n}$. The optimization is performed across all stacks simultaneously, where the earlier viscous blocks learn coarse dynamics, and as \( \gamma_i \) decreases, the subsequent stacks learn high-gradient, entropy-consistent corrections.

\textbf{Problem:} Given the quasilinear hyperbolic PDE~\eqref{eq:system} on the domain $\Lambda$ and its parabolic relaxation~\eqref{eq:parabolic}, together with noisy measurements on $\Gamma$, the objective is to reconstruct the entropic solution $u$ over $\Lambda$ using a VSR-PINN.  
In the baseline VSR-PINN, all stacks are trained simultaneously with a fixed physics weight $\lambda = 1$ and a static set of collocation points $\mathcal{D}_{\text{phy}}$, while the viscosity schedule $\{\gamma_i\}_{i=0}^{n}$ provides an incremental refinement from the parabolic to the hyperbolic model.  

This paper studies the effect of explicitly controlling the curriculum learning of  
(i) the evolution of the physics weight $\lambda$, and  
(ii) the distribution of collocation points $\mathcal{D}_{\text{phy}}$, on the accuracy and robustness of the VSR-PINN.

The following sections investigate three different curriculum strategies for $\lambda$ and $\mathcal{D}_{\text{phy}}$, and evaluate how these choices influence the training dynamics and reconstruction accuracy of the VSR-PINN when solving discontinuous hyperbolic PDEs.

\section{Methodology}

This section presents the proposed framework for improving the VSR-PINN. It is composed of three curriculum techniques: PD optimization, causality, and an adaptive sampling scheme, which are modified and integrated into the architecture. Before explaining each component, the overall training workflow that combines PD optimization with adaptive sampling is provided in Algorithm~\ref{alg:PM-sample}.

\begin{algorithm}
\caption{Baseline algorithm for~\eqref{eq:training}}
\label{alg:PM-sample}
    \begin{algorithmic}[1]
    \FOR{$k = 0$ to $N_{\text{epoch}}$}
        \IF{$k>0$ and $k \bmod N_{\text{resample}}=0$}
        \STATE Do resample 
        \ENDIF
        \STATE Primal update
        \IF{$k \bmod N_\lambda = 0$ and $k > 0$}
            \STATE Dual update
        \ENDIF
    \ENDFOR

\end{algorithmic}
\end{algorithm}

\subsection{Stack-wise Primal-Dual (PD) optimization}

The VSR-PINN is augmented with a PD optimization, which serves as a stack-wise curriculum by automatically learning the physics loss weights.  
In this framework, the fixed original Lagrange multiplier $\lambda$ is replaced by a nonnegative vector of dual variables 
$\boldsymbol{\lambda} = (\lambda_0, \ldots, \lambda_n)$, one for each stack.  
Gradient descent on the network parameters is alternated with projected gradient ascent on $\boldsymbol{\lambda}$.  

From a curriculum perspective, it is suggested to begin with $\boldsymbol{\lambda} = \mathbf{0}$, and the PD scheme adaptively increases $\lambda_i$ only when the current stack’s PDE residual remains large.  
As presented in Algorithm \ref{alg:PD}, this mechanism acts as a \emph{gating} strategy, drawing attention to stages that require physics correction and preventing the optimizer from over-committing to already minimized parts of the stack. 

This behavior aligns with the interpretation in \citep{barreau_accuracy}, where increasing $\lambda$ moves the system along an entropy-increasing training path that avoids poor local minima of the physics term, thereby implementing an effective curriculum optimization rather than relying on manual adjustments. 

\begin{algorithm}
\caption{Stack-wise Primal-Dual (PD) algorithm for~\eqref{eq:training}}
\label{alg:PD}
\textbf{Input:} $\eta_\Theta$, $\eta_\lambda$, $N_{\mathrm{epoch}}$, $N_{\mathrm{\lambda}}$, $N_{\mathrm{resample}}$, $n$.\\
    \begin{algorithmic}[1]
    \STATE Initialize $\Theta_0 \sim \mathcal{G}$ randomly; $\{\gamma_i\}_{i=0}^{n}$; $\boldsymbol{\lambda}=0$.
    \FOR{$k = 0$ to $N_{\text{epoch}}$}
        \IF{$k>0$ and $k \bmod N_{\text{resample}}=0$}
            \IF{\texttt{stopping{\textunderscore}criterium}(k) $<$ $\delta$}
            \STATE{\textbf{break}}
            \ENDIF
        \STATE Do resample
        \ENDIF
        \STATE Primal step:        
        $\Theta_{k+1} 
        = \Theta_{k} 
        - \eta_{\Theta} 
        \nabla_{\Theta}         \mathcal{L}_{\boldsymbol{{\lambda}_k}}\left(
        \Theta_{k}, 
        \gamma, 
        \mathcal{D}_{\text{phy}}
        \right)$
        
        \IF{$k \bmod N_\lambda = 0$ and $k > 0$}
            \STATE Dual step:
          $ \boldsymbol{\lambda}_{k+1}
            = \boldsymbol{\lambda}_{k} 
            + \eta_{\lambda} 
            \nabla_{\!\boldsymbol{\lambda}_{k}} 
            \mathcal{L}\!\left(
            \Theta_{k}, 
            \mathcal{D}_{\text{phy}}
            \right)$
           \ELSE
        \STATE{$\boldsymbol{\lambda}_{k+1} \leftarrow \boldsymbol{\lambda}_{k}$}
        \ENDIF        
        \ENDFOR
    
    \textbf{Return:} $\Theta_{k}$, $\boldsymbol{\lambda}_{k}$

\end{algorithmic}
\end{algorithm}

Overall, the pair $(\gamma_i, \lambda_i)$ defines a two-dimensional curriculum: the viscosity schedule controls conditioning and capacity, while the dual variables assign greater emphasis to stacks with larger residuals.

The stopping criterion is triggered when the network output stabilizes on a held-out set such as $\mathcal{D}_{\text{test}}$. In other words, the mean squared change of the prediction over the last $N_{\text{resample}}$ epochs, relative to that from $N_{\text{resample}}$ steps earlier, is computed as
\begin{multline}
e_{k}(i) = \frac{1}{|\mathcal{D}_{\text{test}}|}
\sum_{(t, x) \in \mathcal{D}_{\text{test}}} 
\left| \hat{u}(\cdot; \Theta_{j}) 
- \hat{u}(\cdot; \Theta_{k - N_{\text{resample}}}) \right|^{2}
\end{multline}
\[
S(k)=\frac{1}{N_{\text{resample}}} \sum_{i = k - N_{\text{resample}}}^{k - 1} e(i),
\]
where the training is stopped if $S(k)$ falls below a specified threshold.

\subsection{Stack-wise causality}

The main goal of the causal training algorithm proposed by \citep{wang_causality} is to reweight the PDE residual in proportion to the inverse exponential of the cumulative residual loss from previous time steps.  
To preserve causality in the vanishing stacked residual PINN, the concept is adapted in two complementary ways:  
(i) a temporal causality mechanism that reweights PDE residuals over time for each stack, and  
(ii) a stack-wise causality mechanism that activates the next stack based on the cumulative gradient norm of the residual loss of the preceding stacks.  
Together, these curricula prevent later stacks from overfitting noisy corrections before the low-fidelity dynamics have stabilized.

For each stack \( i \), the spatially averaged squared residual at each time \( t_{\ell} \) is
\[
\mathcal{L}\big(\{t_{\ell}\}_{\ell=1}^{N_{t}}; \hat{u}^{(i)}, \gamma_{i}\big)
= \frac{1}{N_{x}} 
\sum_{m=1}^{N_{x}}
\left| r_{\gamma_{i}}\!\left(t_{\ell}, x_{m}; \hat{u}^{(i)}\right) \right|^{2}.
\]

The running temporal average is then computed to obtain the temporal weights $W$ and the temporally weighted physics loss $\mathcal{L}_{\mathrm{tc}}$ of stack $i$:
\[
W\!\left(t_{\ell}; \hat{u}^{(i)}, \gamma_{i}\right)
= \exp\!\left(-\kappa \, \mathcal{L}\left(t_{\ell}; \hat{u}^{(i)}, \gamma_{i}\right)\right),
\]
\[
\mathcal{L}_{\mathrm{tc}}\!\left(\hat{u}^{(i)}, \gamma_{i}\right)
= \frac{1}{N_{t}}
\sum_{\ell=1}^{N_{t}}
W\!\left(t_{\ell}; \hat{u}^{(i)}, \gamma_{i}\right)\,
\mathcal{L}\!\left(t_{\ell}; \hat{u}^{(i)}, \gamma_{i}\right),
\]
where $\kappa > 0$ is a hyperparameter that controls the strength of the temporal causality weighting. At training epoch $k$, only the first $D_k$ stacks are active:
\[
\mathcal{S}_k = \{0, 1, \ldots, D_k - 1\}, 
\quad 1 \leq D_k \leq n + 1,
\]
where only $\left\{ \mathcal{L}_{\text{tc}}(\hat{u}^{(i)}, \gamma_i) \right\}_{i \in \mathcal{S}_k}$ are evaluated and backpropagated. Moreover, $D_0 = 1$ represents the baseline PINN, which is the only active block at the beginning of training.  
For each active stack $i\in \mathcal{S}_k$, the gradient norm over all trainable parameters is computed as
\begin{equation}
\label{eq:grad_norm}
g_i^k = 
\left( 
\sum_{\theta_i} 
\left\| 
\nabla_{\theta_i} \mathcal{L}_{\text{tc}}(\hat{u}^{(i)}, \gamma_i) 
\right\|_2^2 
\right)^{1/2}.    
\end{equation}

Subsequently, the causal weight assigned to stack $i$ at epoch $k$ is obtained as
\[
\omega_i^k = 
\exp\left(
    -\varepsilon \, 
    \frac{1}{n + 1} 
    \sum_{j = 0}^{i} g_j^k
\right),
\]
where $\omega_i^k \in (0, 1]$.  
When the gradient norm up to the \( i^{\text{th}} \) stack is large, the training of deeper stacks is deferred until earlier ones reach a certain unlocking threshold \( \vartheta \in (0, 1) \) for \( \tau \) consecutive epochs.  
The parameter \( \varepsilon \) denotes the causality coefficient, which governs the gating mechanism.  
In other words, smaller values of \( \varepsilon \) lead to larger \( \omega_i \) values, thereby accelerating progression to subsequent stacks.  
Conversely, large values of \( \varepsilon \) hinder advancement to the next stack unless the current stacks’ gradients become negligible.

Since both $W$ and $\omega_i^k$ depends on the trainable parameters \( \boldsymbol{\theta} \), 
the \texttt{.detach()} operation in PyTorch or \texttt{tf.stop\_gradient} in TensorFlow 
is used to prevent gradient backpropagation through their computation. Mathematically, the operator $\operatorname{sg}(x) = x$ is defined for the forward pass, while the backward pass satisfies $\nabla_{x} \operatorname{sg}(x) = 0$. That is, $\operatorname{sg}(\cdot)$ forwards the value but blocks its gradient.
Finally, training terminates when the final stack \( n \) satisfies 
\( \omega_n^k \geq \vartheta \) for \( \tau \) patience epochs.

\subsection{Adaptive sampling}

The residual-based adaptive refinement (RAR) method was originally introduced in \citep{Lu_deepxde}, which aims to sample more points from regions where PDE residuals are high, thereby improving the overall distribution of residual points \citep{WU_adaptive}.  
In this work, the approach is adapted to respect the multi-fidelity structure of the VSR-PINN by evaluating residuals at every stacked block rather than only at the final stage.

\begin{algorithm}
\caption{Residual-based adaptive refinement (RAR)}
\label{alg:RAR}
\textbf{Input:} $\mathcal{D}_\text{phy}$, $N_\text{epoch}$, $N_{\text{resample}}$, $m_\text{new}$, $n$. \\
    \begin{algorithmic}[1]
    \FOR{$\mathrm{k}=0$ to $N_{\text{epoch}}$}
        \IF{$k \bmod N_\text{resample}=0$}
            \STATE Select $N_{\text{phy}}$ random points uniformly in $\Lambda$.
            
             \FOR{$i = 0$ to $n$}
                \STATE \textbf{compute} $\left\{ r_{\gamma_i}(j) \right\}_{j=1}^{N_{\text{phy}}}$
                \STATE $q\leftarrow\lfloor m_{\mathrm{new}}/n\rfloor$, $m_r\leftarrow n_{\mathrm{new}}-nq$, set $q_i\leftarrow q+\mathbb{I}[i<m_r]$
                \STATE \textbf{Define} 
               $J_{i} := \operatorname*{arg\,top}^{q_{i}}_{j \in \{1, \ldots, N_{\text{phy}}\}} \left| r_{\gamma_{i}}(j) \right|.$              
             \ENDFOR
         \STATE $J := \bigcup_{i=0}^{n} J_{i}.$
         \STATE $\mathcal{D}_{\text{phy}} \leftarrow \mathcal{D}_{\text{phy}} \cup \{(x_{j}, t_{j}) : j \in J\}$
        \ENDIF
    \ENDFOR
\STATE \textbf{Return:} Updated $\mathcal{D}_\text{phy}$
\end{algorithmic}
\end{algorithm}

\section{Numerical results and discussion}

This section aims to compare the curriculum strategies integrated into VSR-PINN \citep{eshkofti2025} for reconstructing the vehicle density along a one-dimensional road segment $[0, L]$ from density measurements \footnote{The code and data will be available upon acceptance of the manuscript.}. Two density measurements are provided at the boundaries, $x \in \{0, L\}$, using loop detectors. The traffic dynamics follow the Lighthill–Whitham–Richards (LWR) conservation law \citep{RichardShock, lighthill1955kinematic}. The state $u(t, x)$ denotes the normalized density, where $u = 0$ corresponds to an empty road and $u = 1$ represents bumper-to-bumper conditions. The flow–density relationship is modeled using the classical Greenshields flux, $f(u) = V_{f} u (1 - u)$, where $V_{f}$ is the free-flow speed \citep{greenshields1935study}. This nonlinear advection flux can produce shocks and other discontinuities in the solution. 

\begin{figure}[!t]
    \centering
    \includegraphics[width=\columnwidth]{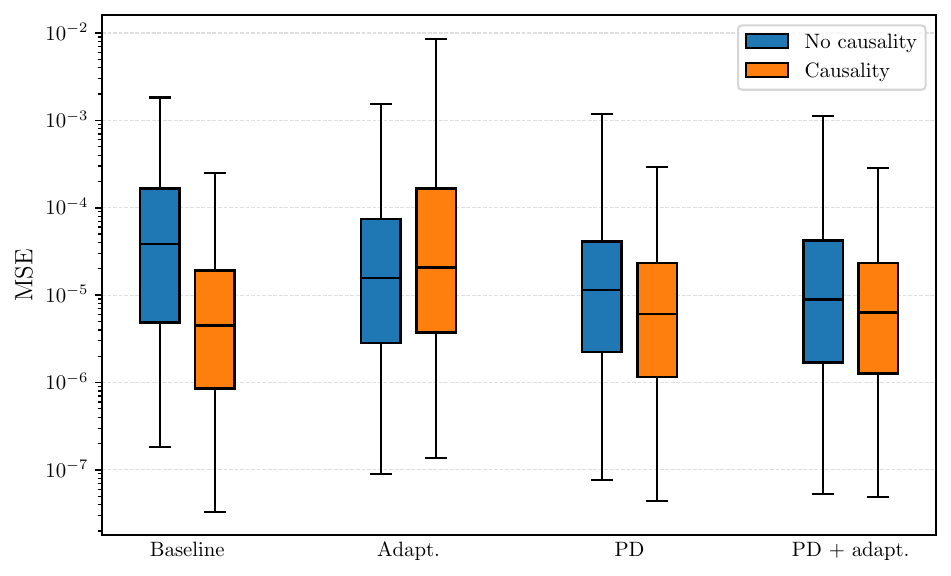}
    \caption{Point-wise MSE distribution for all VSR-PINN variants over three independent runs, comparing causal and non-causal training.}
    \label{fig:boxplot}
\end{figure}

The initial viscosity coefficient is set to $\gamma_{\text{init}} = 0.1$. Since the effect of stacked layers has already been investigated in \citep{eshkofti2025}, a vanilla PINN and three stacked residual layers are considered for all cases. The Godunov simulation from~\eqref{eq:system} is employed to generate the density measurements used in $\mathcal{L}_{\text{data}}$. The vanilla PINN consists of three hidden layers, each with 40 neurons, while each residual stack comprises four hidden layers with 50 neurons. In the superlinear schedule described in \eqref{eq:viscosity}, the parameter is set to $p = 0.5$, and the algorithm runs for up to 20,000 iterations unless terminated earlier by the early stopping criterion. In the causality variants, the temporal causality hyperparameter is set to $\kappa = 10$, while the stack-wise causality factor starts from $\epsilon = 0.1$ with an annealing factor of $5$. That is, as the algorithm progresses to deeper stacks, the factor is multiplied by the annealing value. To systematically analyze the contribution of each curriculum learning method, eight distinct scenarios were considered. The first four correspond to non-causal training and include: (1) the baseline VSR-PINN \citep{eshkofti2025}, (2) the VSR-PINN employing residual-based adaptive sampling, (3) the VSR-PINN using stack-wise PD optimization, and (4) the VSR-PINN combining both mechanisms. The remaining four scenarios replicate the same configurations under causal training, where respecting temporal causality and stack activation occurs according to the proposed causality progression criterion. The relative $\mathcal{L}^2$ error for all these scenarios are listed in \ref{tab:l2_summary}. In addition, Fig.~\ref{fig:boxplot} summarizes the point-wise MSE distributions across three independent implementations of each VSR-PINN configuration. For each configuration, the model is retrained three times with different random initializations. A clear and consistent pattern emerges: for all four variants, the causal versions yield lower median MSEs and a reduced spread compared to their non-causal counterparts. This effect is most pronounced in the Baseline and PD architectures, where causality shifts the entire error distribution downward by roughly one order of magnitude, indicating both improved accuracy and more stable training. It should be noted that, in Fig.~\ref{fig:boxplot}, the causal VSR-PINN with adaptive sampling does not reduce the error further compared to the non-causal version. This can be explained by the fact that, while causality requires early stacks to stabilize before deeper stacks are activated, adaptive sampling continuously injects new high-residual points into the active stack. This keeps its gradients large and repeatedly delays the activation and convergence of the deeper stacks.

\begin{figure}[!t]
     \centering
     \begin{subfigure}[b]{0.5\textwidth}
         \centering
         \includegraphics[width=0.9\textwidth]{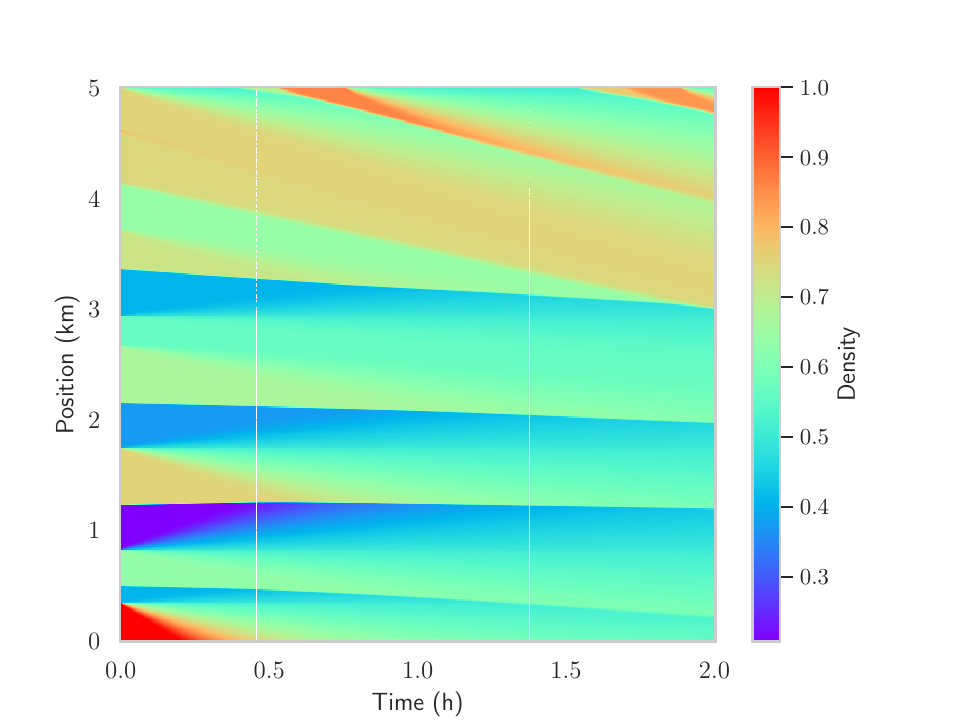}
         \caption{$u$}
         \label{fig:true}
     \end{subfigure}
     \begin{subfigure}[b]{0.5\textwidth}
         \centering
         \includegraphics[width=0.9\textwidth]{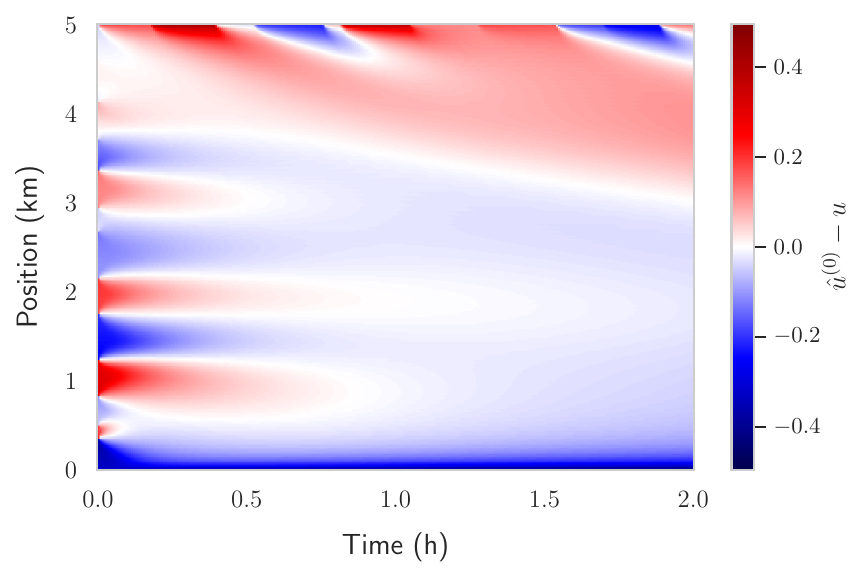}
         \caption{$\hat{u}^{(0)}-u$}
         \label{fig:PINN}
     \end{subfigure}
     \begin{subfigure}[b]{0.5\textwidth}
         \centering
         \includegraphics[width=0.9\textwidth]{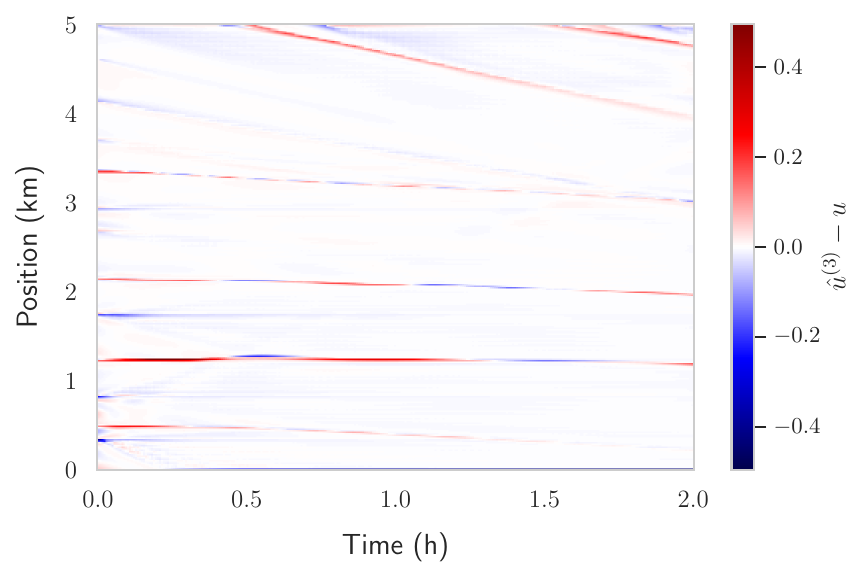}
         \caption{$\hat{u}^{(3)}-u$}
         \label{fig:stack_three}
     \end{subfigure}
    \caption{Reconstruction of the traffic state using the VSR-PINN equipped with causality and PD.}
    \label{fig:numerical_simulation}
\end{figure}

\begin{figure}[!t]
    \centering
    \includegraphics[width=\columnwidth]{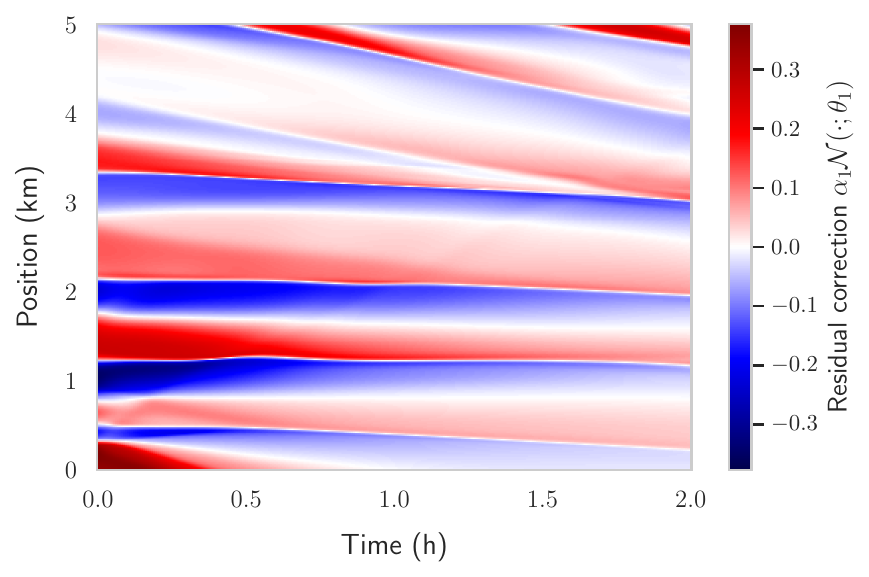}
    \caption{Distribution of the residual block $\alpha_1 \mathcal{N}(\cdot; \theta_1)$..}
    \label{fig:residual}
\end{figure}

Moreover as seen in Fig.~\ref{fig:numerical_simulation}, the reconstruction error of baseline PINN $\hat{u}^{(0)}-u$ shows large deviations near shock waves due to the limitations imposed by smooth parabolic regularization. However, the reconstruction error of the final stack expressed as $\hat{u}^{(3)}-u$ and shoiw in Fig.~\ref{fig:stack_three} reduces the error significantly. Notably, Fig.~\ref{fig:residual} shows that strong positive and negative bands appear precisely along the shock trajectories and in regions where the density gradient is large. This indicates that the first residual block does not relearn the full solution; rather, it serves as a targeted, physics-aware refinement that selectively amplifies or attenuates the baseline prediction near discontinuities.

\begin{table}
\centering
    \renewcommand{\arraystretch}{1.2}
\caption{Comparison of relative $\mathcal{L}^2$ error $(\times 10^{-2})$ for VSR-PINN configurations.}
\label{tab:l2_summary}
\renewcommand{\arraystretch}{1.2}
\setlength{\tabcolsep}{5pt}
\begin{tabular}{l||cccc}
\hline
\# Training  & Baseline & Adaptive & PD & Adaptive + PD \\
\hline\hline
Non-causal & 5.13 & 4.71 & 4.63 & 4.49 \\
Causal     & 3.84 & 4.50 & 3.78 & 4.13 \\
\hline
\end{tabular}
\end{table}

\section{Conclusion}
This paper advances the VSR-PINN framework for reconstructing hyperbolic PDEs, with a particular focus on traffic flow dynamics governed by the LWR conservation law. The proposed architecture integrates three complementary curriculum learning strategies: stack-wise PD optimization, causality progression, and adaptive sampling, within a vanishing viscosity training scheme. The inclusion of PD optimization adaptively balances the competing data and physics losses across stacked layers without manual tuning, while the causality mechanism enforces the progressive activation of residual stacks and ensures stable viscous to inviscid learning. Moreover, the adaptive sampling strategy effectively identifies and concentrates collocation points in regions with high residuals. Although the VSR-PINN demonstrates superior reconstruction accuracy and faster convergence compared with the baseline PINN, the combination of these curriculum strategies, particularly causality, further enhances its robustness. Future work will focus on extending the approach to multidimensional systems and applying it to real world sensor data for large scale traffic networks.

\bibliography{ifacconf}             

\end{document}